\pdfoutput=1

\documentclass[10pt,twocolumn,letterpaper]{article}

\usepackage{cvpr}              

\usepackage{graphicx}
\usepackage{amsmath}
\usepackage{amssymb}
\usepackage{booktabs}
\usepackage{algorithm,algpseudocode}

\usepackage[d]{esvect}

\usepackage[pagebackref,breaklinks,colorlinks]{hyperref}

\usepackage[capitalize]{cleveref}
\crefname{section}{Sec.}{Secs.}
\Crefname{section}{Section}{Sections}
\Crefname{table}{Table}{Tables}
\crefname{table}{Tab.}{Tabs.}

\def\cvprPaperID{*****} 
\def\confName{CVPR}
\def\confYear{2022}

\begin{document}

\title{Point Cloud Recognition with Position-to-Structure Attention Transformers}

\author{Zheng Ding\\
UC San Diego\\
\and
James Hou\\
The Bishop's School\\
\and
Zhuowen Tu\\
UC San Diego
}
\maketitle

\begin{abstract}
In this paper, we present Position-to-Structure Attention Transformers (PS-Former), a Transformer-based algorithm for 3D point cloud recognition. PS-Former deals with the challenge in 3D point cloud representation where points are not positioned in a fixed grid structure and have limited feature description (only 3D coordinates ($x, y, z$) for scattered points). Existing Transformer-based architectures in this domain often require a pre-specified feature engineering step to extract point features. Here, we introduce two new aspects in PS-Former: 1) a learnable condensation layer that performs point downsampling and feature extraction; and 2) a Position-to-Structure Attention mechanism that recursively enriches the structural information with the position attention branch. Compared with the competing methods, while being generic with less heuristics feature designs, PS-Former demonstrates competitive experimental results on three 3D point cloud tasks including classification, part segmentation, and scene segmentation.
\end{abstract}

\section{Introduction}
\label{sec:intro}


3D point cloud recognition is an active research area in computer vision that has gained steady progress in the past years \cite{qi2017pointnet,li2018pointcnn,qi2017pointnet++,xie2018attentional,guo2020pct}. The availability of large-scale 3D datasets \cite{wu20153d} and real-world applications \cite{hu2020randla} in autonomous driving \cite{chen2017multi,meyer2019lasernet}, computer graphics \cite{mo2019structurenet}, and 3D scene understanding \cite{jaritz2019multi} make the task of 3D point cloud recognition increasingly important.

\begin{figure}[!htp]

\begin{center}

\scalebox{0.57}{
\begin{tabular}{c}
\includegraphics[width=0.8\textwidth]{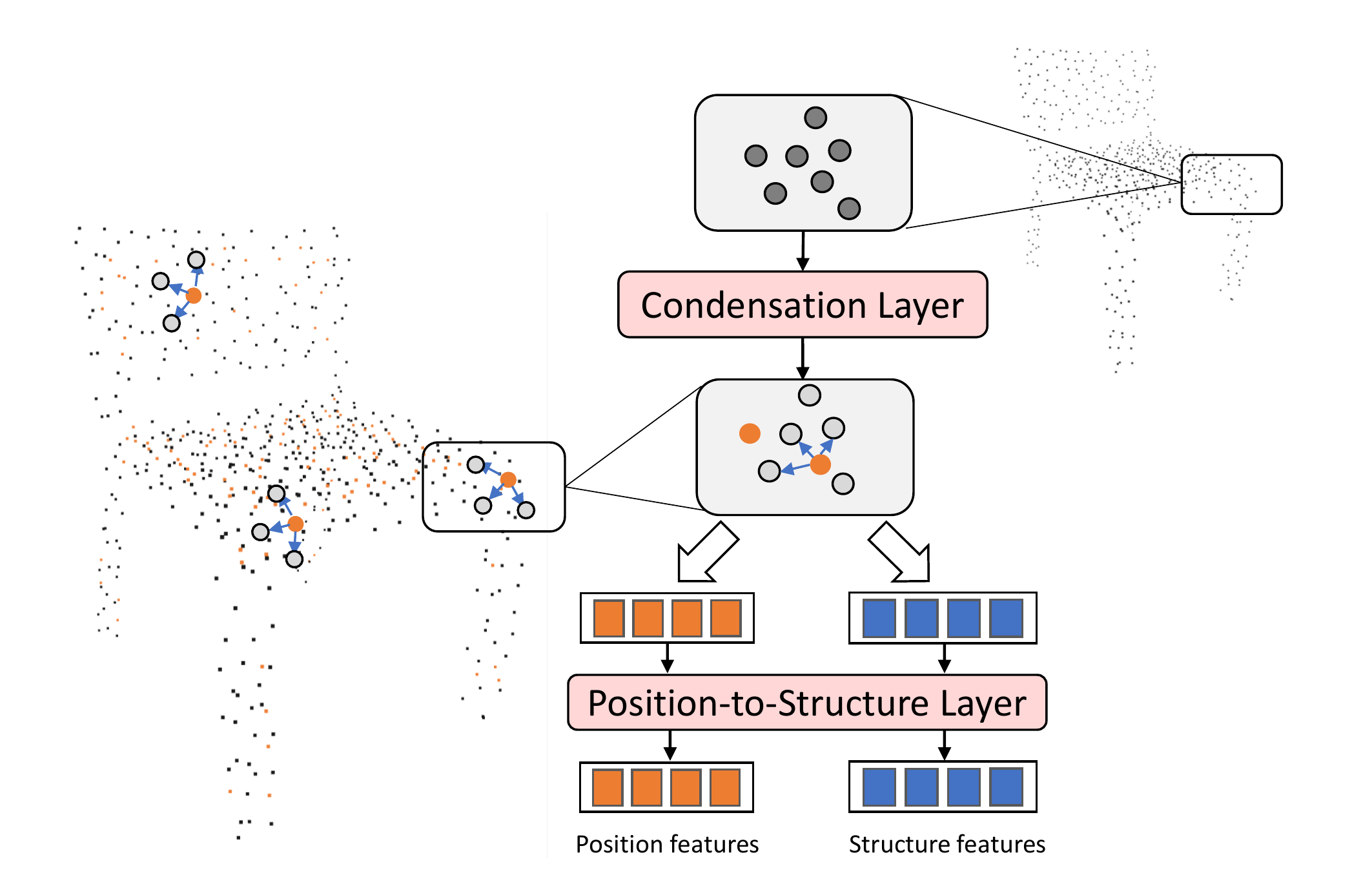}
\end{tabular}
}
\caption{\small \textbf{Overview of our proposed method}: Our PS-Former learns rich feature descriptions together with robust point local graphs. Our PS-Former shows more effectiveness in learning 3D point cloud representations than vanilla self-attention and cross-attention mechanisms in the standard Transformers, as demonstrated in the ablation study (Section \ref{subsec:ablation-psa}). Left: Illustration of local structure relations (local graphs of points and their neighboring points). Right: PS-Former Pipeline. It first extracts the structure relations (local graphs) of individual points via a learnable Condensation Layer. Then, a Position-to-Structure Attention mechanism is applied to enrich the structure features from the position features in a recursive fashion.}
\label{fig:tease}
\end{center}
\vspace{-10mm}
\end{figure}


Amongst recent 3D shape representations for deep-learning based recognition tasks, including meshes \cite{gkioxari2019mesh}, voxels (or volumetric grid) \cite{wu2017marrnet}, and implicit functions \cite{chen2019learning},
the point cloud representation \cite{qi2017pointnet} remains a viable choice to represent 3D shapes due to its flexibility and effectiveness in computation and modeling. However, adopting the point cloud representation in downstream tasks poses some special challenges: 1) unlike the voxel-based volumetric representation that has a fixed grid-structure where 3D convolutions \cite{wu2017marrnet} can be readily applied, point clouds are basically collections of scattered points that hold no order; 2) each sample point carries only the 3D coordinate information ($x, y, z$) without rich explicit feature descriptions. A good understanding about the overall shape, as well as the object parts represented by point clouds depends on extracting the ``correct'' and ``informative'' features of the individual points through their relations with the neighboring points (context).

\textbf{The main challenge: a chicken-and-egg problem}.
Point clouds come in as scattered points without known connections and relations. Point cloud recognition is a chicken-and-egg problem: a rich feature description benefits from the robust extraction of the point structure relations (local graphs), whereas creating a reliable local graph also depends on an informative feature description for the points. In a nutshell, the local graph building and feature extraction processes are tightly coupled in 3D point cloud recognition, which is the central issue we are combating here.


Transformers \cite{vaswani2017attention} are emerging machine learning models that have undergone exploding development in natural language processing \cite{devlin2019bert} and computer vision \cite{carion2020end,dosovitskiy2021image}. Unlike Convolutional Neural Networks \cite{lecun1989backpropagation} that operate in the fixed image lattice space, the attention mechanism in the Transformers \cite{vaswani2017attention} include the positional embeddings in the individual tokens that themselves are already orderless. This makes Transformers a viable representation and computation framework for 3D point cloud recognition. 
Using the vanilla Transformers on point clouds \cite{vaswani2017attention,dosovitskiy2021image} under the standard self-attention mechanism, however, leads to a sub-optimal solution (see ablation study in section \ref{subsec:ablation-psa}). The weighted sum mechanism learns an average of the tokens which are not ideal for structure extraction. Existing works like PCT\cite{guo2020pct} have a special non-learnable feature engineering step that pre-extracts features for each sampled point. 

Given these challenges, we propose in this paper a new point cloud recognition method, \textbf{P}osition-to-\textbf{S}tructure Attention TransFormer (\textbf{PS-Former}), that consists of two interesting properties: 
\begin{enumerate}
    \item A learnable \textbf{Condensation Layer} that performs point cloud downsampling and feature extraction automatically. Unlike the Transformer-based PCT approach \cite{guo2020pct}, where a fixed strategy using farthest point sampling and a feature engineering process using KNN grouping are adopted, we extract structural features by utilizing the internal self-attention matrix for computing the point relations.
    \item A \textbf{Position-to-Structure Attention} mechanism that recursively enriches the structure information using the position attention branch. This is different from the standard cross-attention mechanism where two working branches cross attend each other in a symmetric way. An illustration can be found in Figure \ref{fig:psa}.
\end{enumerate}

We conduct experiments on three main point cloud recognition tasks including ModelNet40 classification\cite{wu20153d}, ShapeNet part segmentation\cite{yi2016scalable}, and 3SDIS scene segmentation\cite{armeni20163d}, to evaluate the effectiveness of our proposed model. ModelNet40 classification requires the recognition of the entire input point cloud while the latter two focus on single point labeling. PS-Former is shown to be able to achieve competitive results when compared with state-of-the-art methods, and improve over the PCT method \cite{guo2020pct} that computes the features before the attention layer by grouping them based on the original 3D space distance.

\section{Related Work}

\textbf{3D Point Cloud Recognition.} Due to the point cloud's non-grid data structure, a number of works have been proposed in the past by first converting the points to a grid data structure e.g. 2D images or 3D voxels\cite{su2015multi,goyal2021revisiting}. The grid data after this pre-processing can be directly learned by CNN-like structure \cite{lecun1989backpropagation}. However, this conversion process from point clouds to volumetric data may lead to information loss e.g. occlusion in the 2D images and resolution bottleneck in the 3D voxels.
The seminal work of PointNet~\cite{qi2017pointnet} chose to perform learning directly on the original point cloud data in which max pooling operation is adopted to retain invariance in the point sets. Since the work of PointNet~\cite{qi2017pointnet}, there has been a wealthy body of methods proposed along this direction. There are methods~\cite{thomas2019kpconv,wu2019pointconv,xu2021paconv} attempting to simulate the convolution process in 2D images whereas other approaches~\cite{wang2018local,xu2020grid} adopt graph convolutional neural networks (GCN) to build connections for the neighboring points for feature extraction.\\
\textbf{Transformer Architecture.} A notable recent development in natural language processing is the invention and widespread adoption of the Transformer architectures \cite{vaswani2017attention,devlin2018bert}.  At the core, Transformers \cite{vaswani2017attention} model the relations among tokens with two attention mechanisms, namely self-attention and cross-attention. Another important component of Transformers is the positional encoding that embeds the position information into the tokens, relaxing the requirement to maintain the order for the input data. Recently, Transformers have also been successfully adopted in image classification \cite{dosovitskiy2021image} and object detection \cite{carion2020end}. Typically, vision transformers adopt similar absolute/relative positional encoding strategies used in language transformers \cite{vaswani2017attention,shaw2018self} to encode grid structures from 2D images. In 3D point cloud tasks, the input point cloud data only contains the 3D coordinates without any texture information, which makes the computation from position to structure feature extraction a challenging task. In point cloud recognition, Transformers were adopted in \cite{xie2018attentional}; PCT~\cite{guo2020pct} applies sampling and grouping operations introduced by PointNet++\cite{qi2017pointnet++} to capture the structure information.

Compared to existing 3D point cloud recognition methods \cite{maturana2015voxnet,li2018pointcnn,wang2019dynamic,xiang2021walk}, our proposed PS-Former model consists of learnable components that are generic and easy to adapt. Versus the competing Transformer based approaches such as PCT \cite{guo2020pct}, PS-Former (1) removes the non-learnable feature engineering stage with a Condensation Layer and (2) incorporates newly designed Position-to-Structure Attention that learns informative features from point positions and their neighboring structures.

\section{Method}

In this section, we present our proposed model, Position-to-Structure Transformers (PS-Former). We first give an overview of our model, followed by a description of the two key components of our model: Position-Structure Attention Layer and Condensation Layer.

\begin{figure*}[!htp]
\begin{center}
\scalebox{1.05}{
\begin{tabular}{cc}
\includegraphics[width=0.9\textwidth]{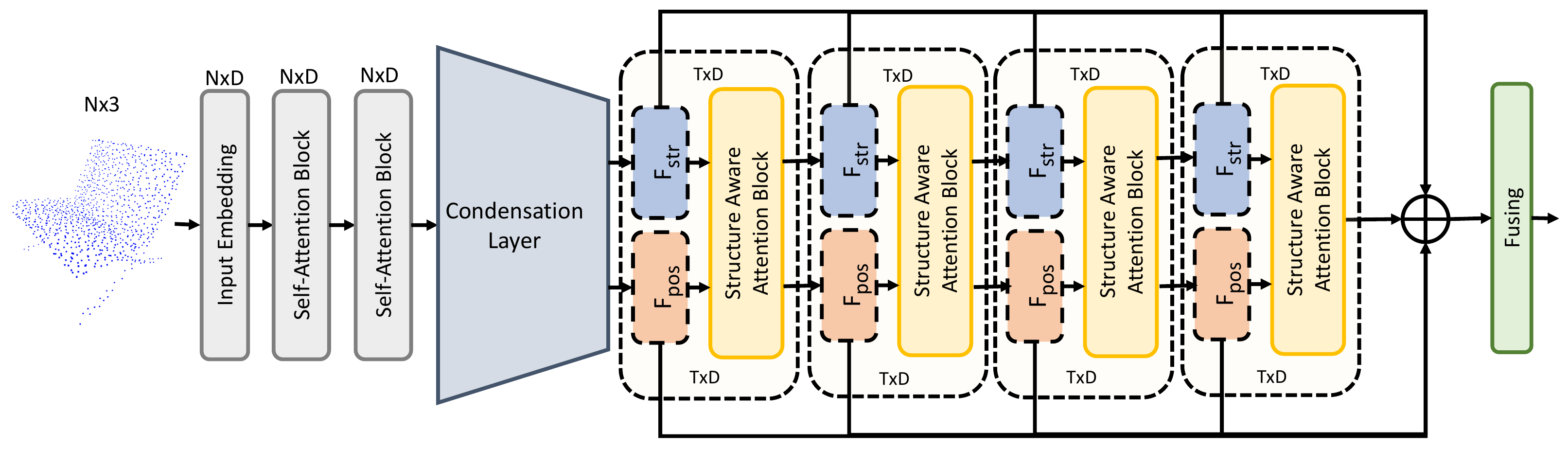}
\end{tabular}
}
\caption{\small \textbf{Schematic illustration of our proposed PS-Former}. The input point cloud first goes through self-attention blocks, and then the point cloud features are downsampled and extracted by a learnable Condensation Layer. Finally, the features are decoupled into a position feature branch and a structure feature branch. A Position-to-Structure Attention mechanism recursively enriches the structure information using the position feature branch, and learns a feature representation for 3D point cloud tasks e.g. classification and segmentation. 
N: Number of points. D: Dimension of features. T: Number of points after the Condensation Layer. $F_{\text{pos}}$: Position feature of each point. $F_{\text{str}}$: Structure feature of each point.}
\label{fig:overview}
\end{center}
\end{figure*}

\subsection{Overview}

Our proposed PS-Former model is a Transformer-based architecture, as seen in Figure \ref{fig:overview}. The overall goal of the PS-Former is to learn an effective representation for downstream tasks such as classification and segmentation. The input is $N\times 3$ coordinate data which will first go through an input embedding module to project the data from the low-dimension space to the high-dimension space. Two self-attention layers are adopted to make points interact with each other to obtain more developed features before condensation. Following are our two key designs: Condensation Layer and Position-to-Structure Attention Layer. The features are concatenated together for further recognition tasks.

\subsection{Attention Mechanism}
The vanilla attention mechanism \cite{vaswani2017attention} is as follows:
\begin{align}
    E = (\frac{QK^T}{\sqrt{C}}),
    A_{ij} =  \frac{e^{E_{ij}}}{\sum_j{e^{E_{ij}}}}
\end{align}
where $Q$ and $K$ are query and key respectively, $C$ is the feature dimensions and $A$ is the final attention matrix. However, in the 3D space of points, a more meaningful way to consider the correlation between points is the euclidean distance instead of the dot product. Thus, we introduce a new attention mechanism based on euclidean distance in our model. We also keep this characteristic in the high dimensional space as well to facilitate an easier learning process. Moreover, we use a similar normalization method as PCT\cite{guo2020pct} and we'll see later in the Condensation Layer (section \ref{method:cond}) it will be a key design for our algorithm. Our final attention equation is as following:
\begin{align}
    E_{ij} = -||Q_{i}-K_{j}||_2^2 \\
    A'_{ij} = \frac{e^{E_{ij}}}{\sum_i{e^{E_{ij}}}}, 
    A_{ij} = \frac{A'_{ij}}{\sum_jA'_{ij}}
\end{align}

\subsection{Condensation Layer}
\label{method:cond}
\begin{figure*}[htbp]
\begin{center}
\scalebox{1}{
\begin{tabular}{c}

\includegraphics[width=0.9\textwidth]{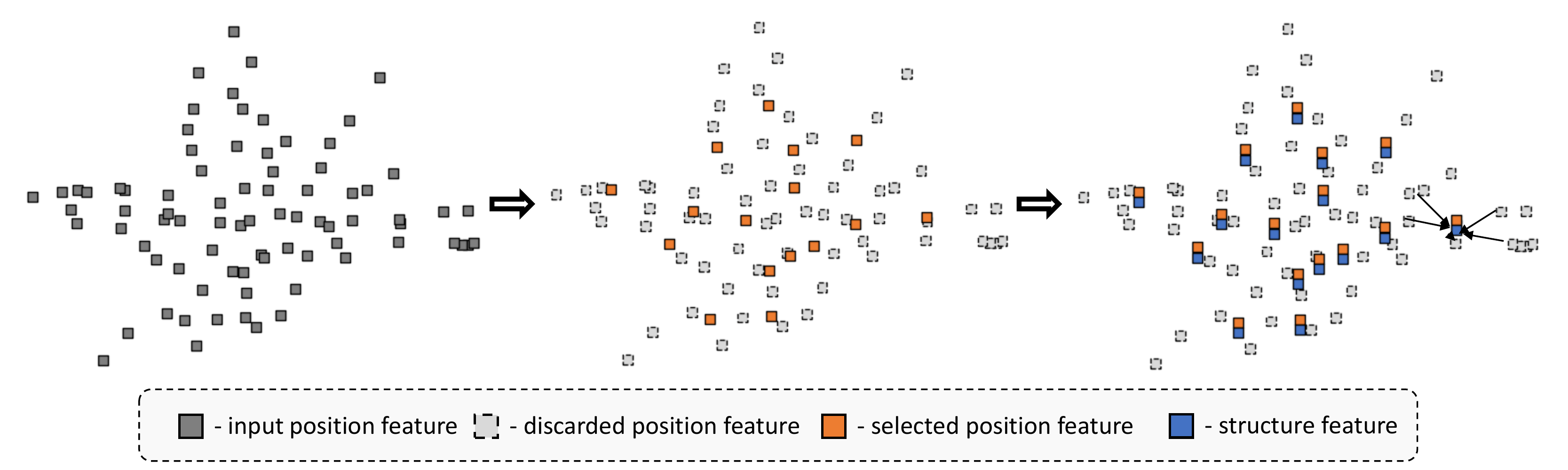}

\end{tabular}
}
\caption{\small \textbf{Condensation Layer:} This is a depiction of the Condensation process, showing a plant's point cloud from ModelNet40 \cite{wu20153d}. Left: In the first stage, we feed a complete point cloud to the Condensation Layer. Middle: Using Algorithm \ref{alg:one}, we select the top-k point features that best represent the point cloud. Right: For each selected position feature, we generate an accompanying structure feature through the surrounding discarded point features.}
\label{fig:condensation_schema}
\end{center}
\end{figure*}

A common challenge in point cloud recognition is exploiting rich local and global structures. Methods such as PointNet++\cite{qi2017pointnet++} and PCT\cite{guo2020pct}  have adopted a combination of sampling---Farthest Point Sampling (FPS)---and grouping methods to tackle this difficulty. 
Our approach extracts the local structures through a two-step process based entirely on attention: sampling and feature construction. The generation of these features through attention create a much richer and more focused local neighborhood context as compared to the more handcrafted approach of PCT and PointNet++. Furthermore, the proposed method proves to be a natural integration into the Transformer network. The Condensation Layer also compresses information for less computational complexity. The schema of the Condensation Layer can be viewed in Figure \ref{fig:condensation_schema}.

We select $T$ points from the point cloud to serve as center points of local neighborhoods. The traditional Farthest Point Sampling strategy does not identify the most important neighborhoods because its main objective, to maximize inter-point distance, does not motivate meaningful center points. Since attention is a learned measure of affinity and an integral part to the Transformer, we sample the center points based on the row sum of the attention matrix. We apply the softmax operation in columns at first, which can be understood as each point's contribution to each other. The row sum can be seen as how much information each point will take. We make the choice of the center points based on this computed row sum. Note that we don't simply choose the top-k points with the largest sum since it will likely cause points to crowd together. The detailed algorithm is shown in Algorithm \ref{alg:one}.

After we get the center points, we use the strategy which will be described in section \ref{section:ptsa} to obtain the structure features for each point from the relative position features. Moreover, the position features are updated using the vanilla attention mechanism.

\begin{algorithm}
\caption{Attention-Based Condensation}\label{alg:one}
\textbf{Input:} Masked Attention: $A\in\mathbb{R}^{N\times N}$, $N$: the number of points, $T$: the number of selected points, $P_n$: input points\\
\textbf{Output:} $P_s$: sampled points
\begin{algorithmic}[1]
\State $e_i \gets \sum_{j=1}^{N}{A_{i,j}}$
\State $id_s \gets {\emptyset}$
\For{$k \gets 1,2...T$}
    \State $r_k \gets \text{argmax}(e)$
    \State $e_i \gets e_i - A_{i, r_k}$
    \State $e_{r_k} \gets -\infty$
    \State Insert $r_k$ into $id_s$
\EndFor
\State $P_s \gets \text{index}(P_n, id_s)$\\
\Return{$P_s$}
\end{algorithmic}
\end{algorithm}

\subsection{Position-to-Structure Attention Layer}
\label{section:ptsa}
\begin{figure}[htbp]
\begin{center}
\scalebox{0.8}{
\begin{tabular}{c}
\includegraphics[width=0.5\textwidth]{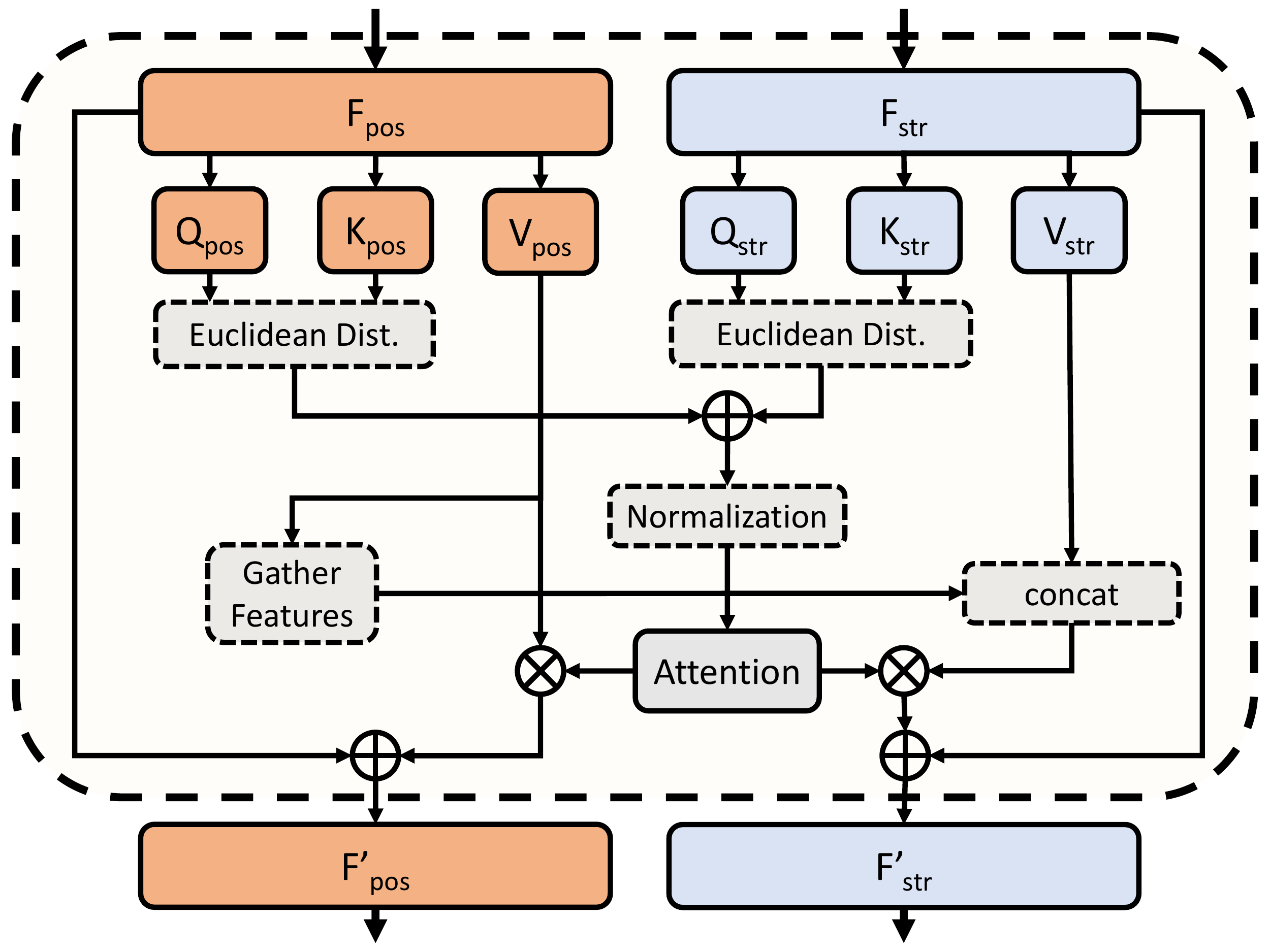}
\end{tabular}}
\caption{\small \textbf{Position-to-Structure Attention Layer:} In the diagram, $F_{\text{pos}}$ and $F_{\text{str}}$ are position feature and structure feature, respectively from the previous Position-to-Structure module. Each input is mapped to a pair of $Q$, $K$ and $V$, labeled str and pos for structure and position. $F'_{\text{pos}}$ and $F'_{\text{str}}$ represent the output position feature and structure feature of the Position-to-Structure Attention Layer to be passed to the next layer.}
\label{fig:psa}
\end{center}
\end{figure}
\begin{figure*}
\begin{center}
\scalebox{0.9}{
\begin{tabular}{ccc}
\includegraphics[width=0.27\textwidth]{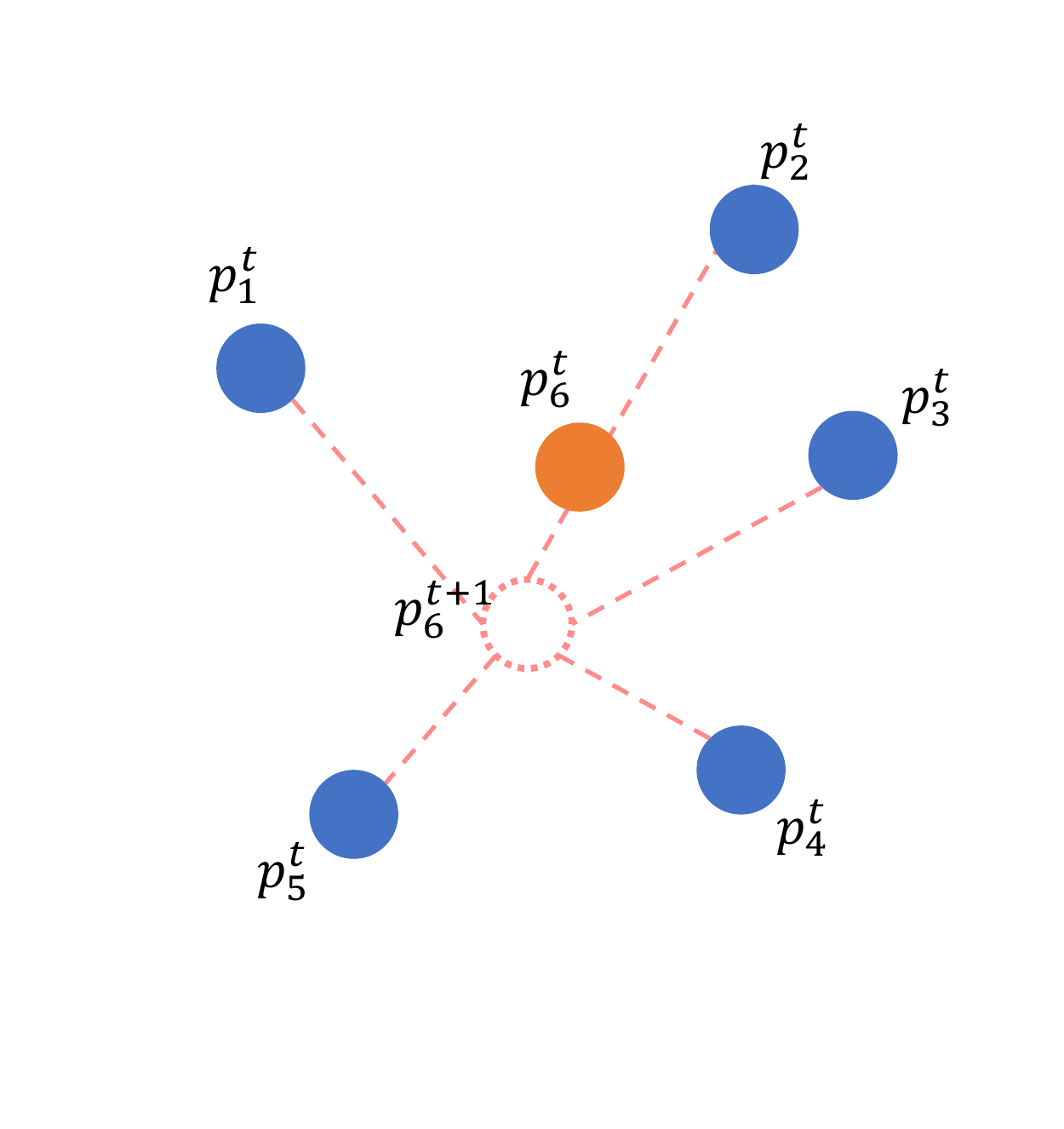} & 
\includegraphics[width=0.27\textwidth]{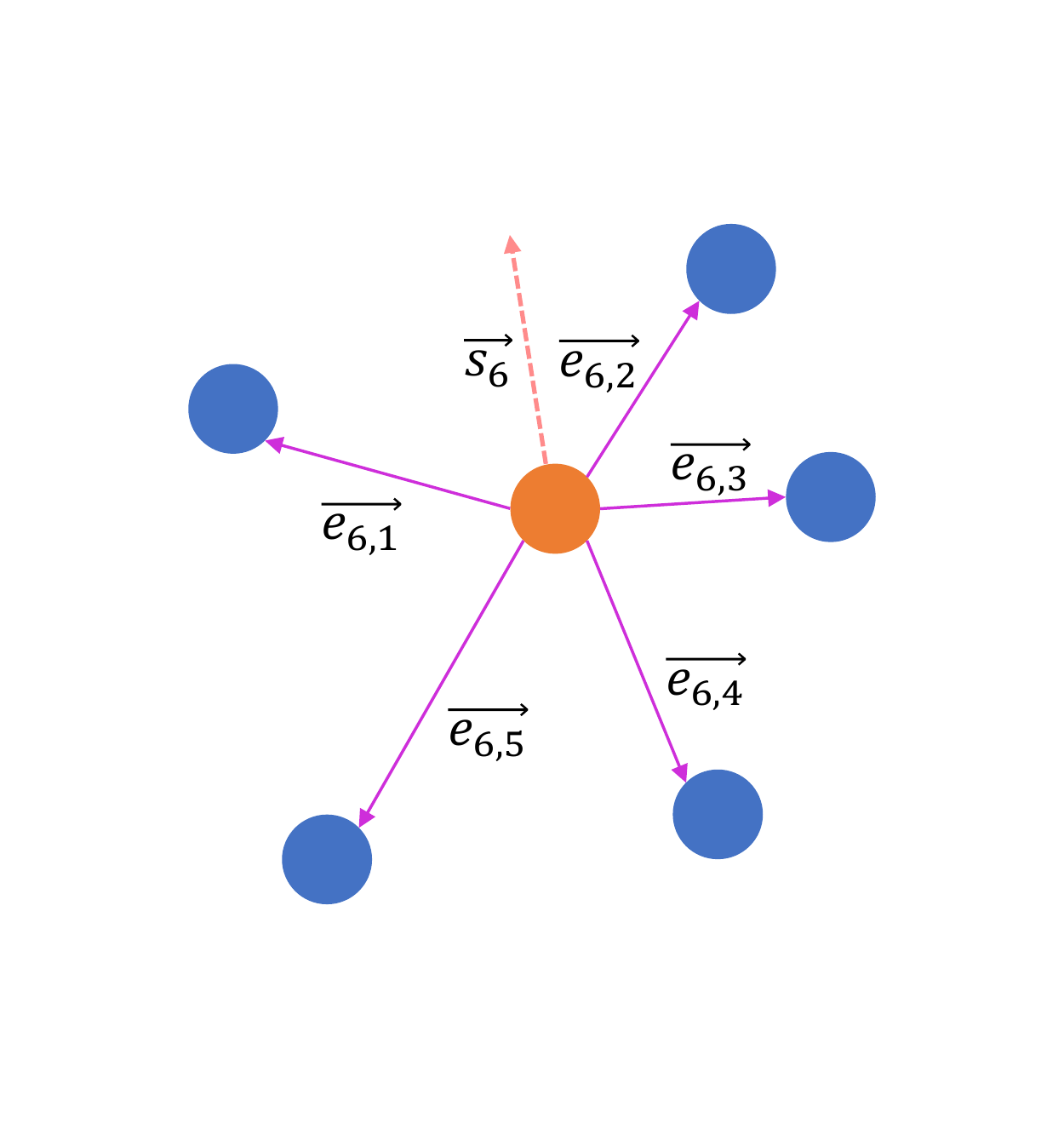} &
\includegraphics[width=0.27\textwidth]{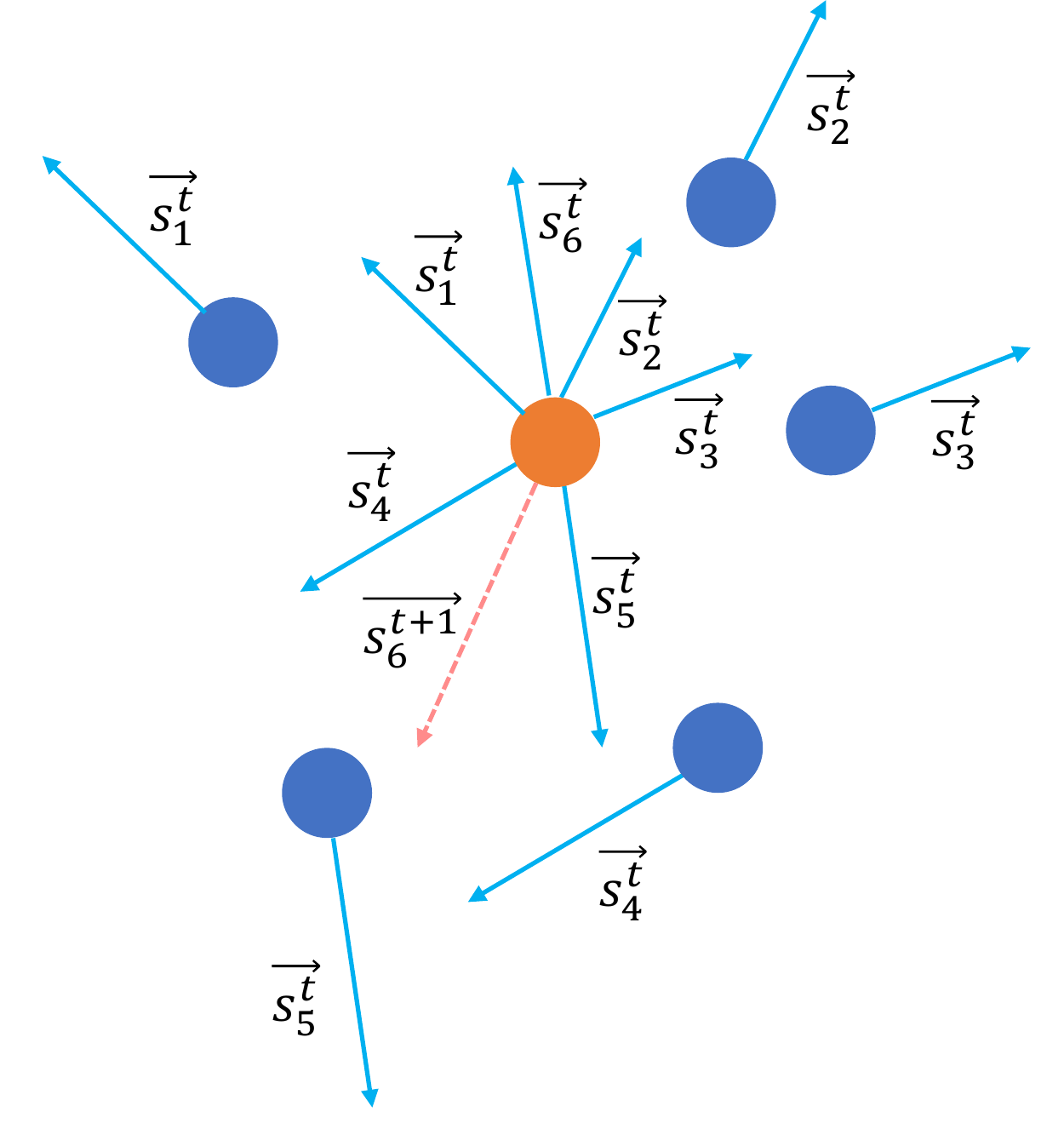} \\
(a) position to position & (b) position to structure & (c) structure to structure
\end{tabular}}
\caption{\textbf{Three feature updating processes in the Position-Structure Attention Layer.} In this case, (a) $p^{t}_{i}$, where $i=1, 2, ..., 5$, position features are used to update $p^{t}_{6}$ to $p^{t+1}_{6}$. (b) relative position features $\protect\vv{{e}_{6,i}}$, where $i=1, 2, ..., 5$, are combined to form structure 
feature ${\protect\vv{s_{6}}}$. (c) structure features $\protect\vv{s^{t}_{i}}$, where $i=1, 2, ..., 5$, are used to update $\protect\vv{s^{t}_{6}}$ to $\protect\vv{s^{t+1}_{6}}$.
}
\label{fig:psa_updating}
\end{center}
\end{figure*}

Our key design is to decouple the position and structure information and allow the position feature to gradually enrich the structure feature. To achieve this, we design the Position-to-Structure Attention Layer as presented in Figure \ref{fig:psa}. The structure information is acquired using the relative position features which can be seen as the neighborhood structure features around one specific point.
In our Position-to-Structure Attention Layer, there are three main processes to update the features which can be seen in Figure \ref{fig:psa_updating}. Each point's position feature is updated through the weighted sum along with its structure feature, while the relative position feature is also used to enrich the structure feature. 

In particular, the attention matrix is computed by both the position features and the structure features as following:
\begin{align}
E_{ij} &= -||Q_i^{\text{pos}}-K_j^{\text{pos}}||_2^2-||Q_{i}^{\text{str}}-K_{j}^{\text{str}}||_2^2,\\ A'_{ij} &= \frac{e^{E_{ij}}}{\sum_i{e^{E_{ij}}}},\ \ A_{ij} = \frac{A'_{ij}}{\sum_jA'_{ij}}
\end{align}

Both the position features and structure features are updated through attention while structure features are also influenced by top-k related points according to the attention matrix.
\begin{align}
F'_{\text{pos}} &= F_{\text{pos}} + \mathcal{K}_1(A\cdot V_{\text{pos}}),\\
F'_{\text{str}} &= F_{\text{str}} + \mathcal{K}_2(A\cdot V_{\text{str}} \circ F_{\text{gather}})
\end{align}
where $\mathcal{K}_1, \mathcal{K}_2$ are mapping functions:  $R^d\rightarrow R^d$, $\circ$ is the concatenate operation, $F_{\text{pos}}$ is the position features for each point, $F_{\text{str}}$ is the structure features for each point and $F_{\text{gather}}$ is the features gathered using the relative position features.

\section{Experiments}

In this section, we evaluate our proposed model in three point cloud recognition tasks: point cloud classification on ModelNet40\cite{wu20153d}, point cloud part segmentation on ShapeNet\cite{yi2016scalable} and indoor scene segmentation on Stanford 3D Indoor Space (S3DIS)\cite{armeni20163d}.

\subsection{Classification on ModelNet40}
\label{subsec:modelnet40-experiment}

\textbf{Dataset and implementation.} A primary assay for point cloud classification, ModelNet40\cite{wu20153d} is a 3D point cloud dataset composed of 12,311 CAD models from 40 different object classes. To maintain consistency with other studies, the same train-test split is used as the given dataset: 9,843 training samples and 2,468 testing samples. In processing the data, we randomly select 1,024 points, as many other methods do. It is important to note that we discard information regarding surface normals and solely use the 3D coordinates. Following PointNet\cite{qi2017pointnet} convention, we augment the training data with random translation, scaling and dropout. While testing, augmentation is omitted and the test set is evaluated without any voting strategy. The model is trained with the Adam optimizer \cite{kingma2014adam} with its learning rate and weight decay both set at 1e-4.

\begin{table}[!htbp]
\begin{center}
\scalebox{0.8}{
\begin{tabular}{l | cccc}
\textbf{Model} & {Input} &  {\# Points} & {Accuracy (\%)}  \\
\hline
{VoxNet}   \cite{maturana2015voxnet}
            & v & - & 85.9   \\
{Subvolume}   \cite{qi2016volumetric}
            & v & - & 89.2   \\
{PointNet}   \cite{qi2017pointnet}
            & pc & 1k & 89.2   \\
{PAT}   \cite{yang2019modeling}
            & pc & 1k & 91.7   \\
{Kd-Net}   \cite{klokov2017escape}
            & pc & 32k & 91.8   \\
{PointNet++}   \cite{qi2017pointnet++}
            & pc & 5k & 91.9   \\
{PointCNN}   \cite{li2018pointcnn}
            & pc & 1k & 92.5   \\
{DGCNN}   \cite{wang2019dynamic}
            & pc & 1k & 92.9   \\
{InterpCNN}   \cite{mao2019interpolated}
            & pc & 1k & 93.0   \\
{GeoCNN}   \cite{lan2019modeling}
            & pc & 1k & 93.4   \\
{PAConv} \cite{xu2021paconv}
            & pc & 1k & 93.6 \\

{RPNet} \cite{ran2021learning}
            & pc & 1k & \textbf{94.1} \\
{CurveNet} \cite{xiang2021walk}
            & pc & 1k & 93.8 \\
\hline
A-SCN   \cite{xie2018attentional}
            & pc & 1k & 89.8   \\
{Point Transformer}   \cite{engel2021point}
            & pc & 1k & 92.8   \\
Cloud Transformers   \cite{mazur2021cloud}
            & pc & 1k & 93.1   \\
PCT   \cite{guo2020pct}
            & pc & 1k & 93.2   \\
{PointTransformer} \cite{zhao2021point} 
            & pc & 1k & 93.7 \\
PS-Former (Ours)
            & pc & 1k & \textbf{93.9} \\ 
\end{tabular}
}
\vspace{1mm}
\caption{\small \textbf{ModelNet40 \cite{wu20153d} Classification:}  Overall accuracy for classification on ModelNet40 with state-of-the-art methods. pc: point cloud, v: voxel. Results in this table are from cited papers and all are without voting strategy.}
\label{tab:modelnet40}
\end{center}
\end{table}

\begin{figure}
\begin{center}
\scalebox{0.9}{
\begin{tabular}{cccc}
\includegraphics[width=0.22\linewidth]{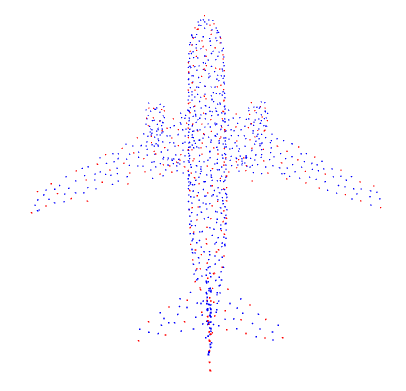}
&
\includegraphics[width=0.22\linewidth]{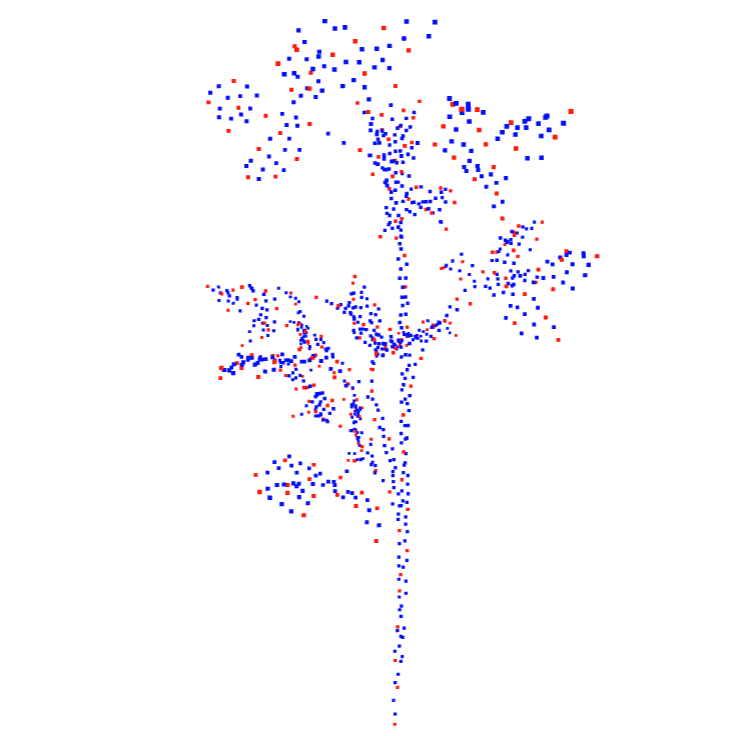}
&
\includegraphics[width=0.22\linewidth]{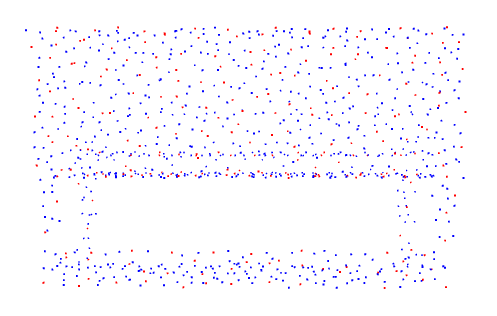}
&
\includegraphics[width=0.22\linewidth]{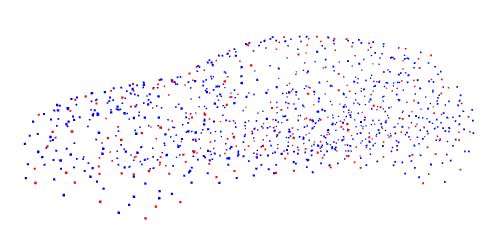}
\\
\includegraphics[width=0.22\linewidth]{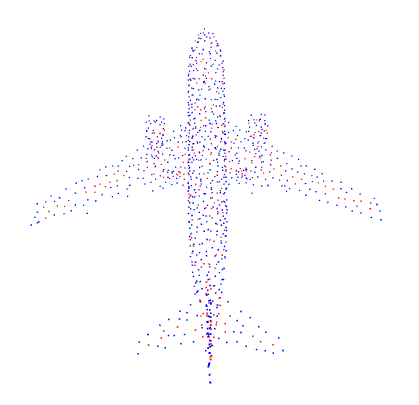}
&
\includegraphics[width=0.22\linewidth]{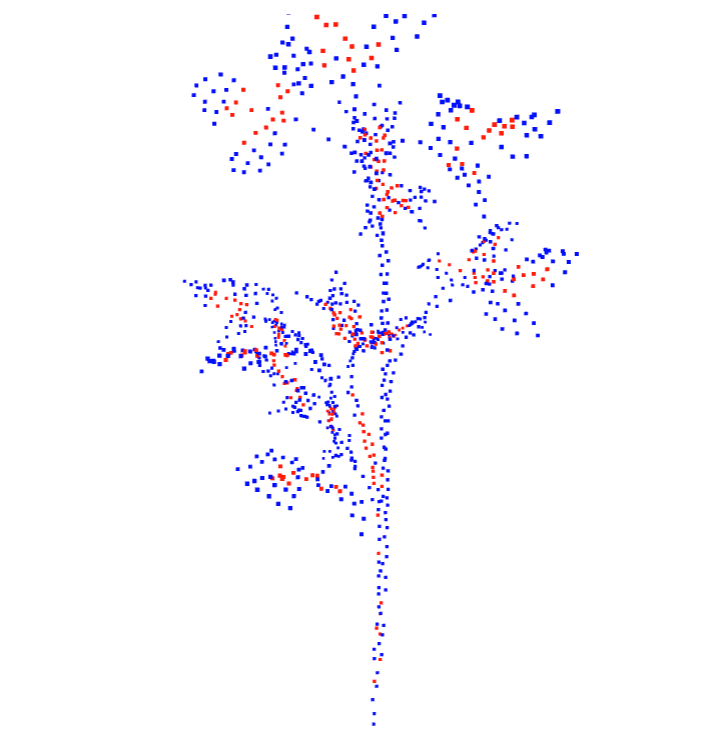}
&
\includegraphics[width=0.22\linewidth]{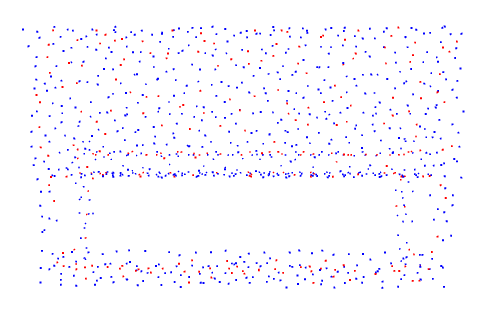}
&
\includegraphics[width=0.22\linewidth]{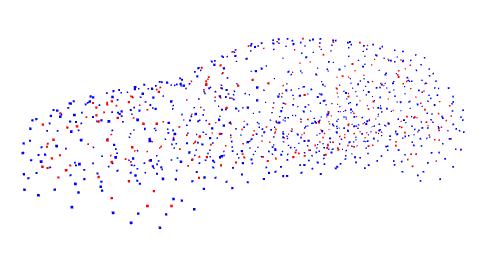}\\
\end{tabular}
}
\caption{\textbf{Visualization of center points.} First Row: Centers extracted by Farthest Point Sampling. Second Row: Centers extracted by our Condensation Layer.  
Red dots represent center points. Plane, plant, chair, and car depicted.}
\label{fig:center-vis}
\end{center}
\end{figure}

\textbf{Results and analysis.} We achieved a competitive result on the ModelNet40 dataset, as seen in Table \ref{tab:modelnet40}. The best overall accuracy achieved by the PS-Former without the assistance of normals or voting ensembles was 93.9\%. For the experiment, we investigated the results of our novel Condensation Layer (shown in Figure \ref{fig:center-vis}). We observe that our Condensation Layer identifies points of interests much more commonly in central areas, while the Farthest Point Sampling (FPS) strategy identifies points in an even distribution in the point cloud. We hypothesize that our Condensation Layer favors more information rich centers that hold high attention values with respect to its neighboring points; the Farthest Point Sampling technique likely produces an even collection of points due to it's objective to maximize point-to-point distance. With intuitively chosen attention-based centers, our model is able to outperform FPS-based methods such as PointNet++ and PCT. Also, with the novel Position-to-Structure Attention and Condensation Layer, our PS-Former outperforms other Transformer-based methods. 

\subsection{Part Segmentation on ShapeNet}

\textbf{Dataset and implementation.} We further evaluate our model on ShapeNet\cite{yi2016scalable} for part segmentation. It contains 16,881 shapes with 14,007 as training samples and the remaining 2,874 as testing samples. The dataset contains 50 object parts in total with each object containing no more than 6 object parts. 2048 points are sampled for each point cloud and all the points are assigned with a part label. To classify each point into the correct label, our Condensation Layer's output token number is set to the input token number which in this case is 2048. After the Position-to-Structure Attention layers, we concatenate the node feature with the whole data feature which is obtained using max pooling as in ModelNet40 classification task. To reduce the computation, in this experiment, we only use the structure feature since this task is more focused on point-wise classification and structure information is more important in this case. We use SGD as our optimizer with the initial learning rate set to 0.1 and the weight decay set to 1e-4. Cosine Annealing Scheduler is used to adjust the learning rate with the minimal learning rate set to 1e-3 \cite{loshchilov2016sgdr}.

\begin{table*}[!htp]
    \centering
	\scalebox{0.75}{
			\begin{tabular} {l|c|cccccccccccccccccccc}
				Method & mIOU & aero & bag & cap & car & chair & ear & guitar & knife & lamp & lap & motor & mug & pistol & rocket & skate & table  \\
				\hline
				PointNet \cite{qi2017pointnet}  & 83.7 & 83.4 & 78.7 & 82.5 & 74.9 & 89.6 & 73.0 & {91.5} & 85.9 & 80.8 & 95.3 &  65.2 & 93.0 & 81.2 & 57.9 & 72.8 & 80.6 \\
				SO-Net \cite{li2018so} & 84.9 & 82.8 & 77.8 & 88.0 & 77.3 & 90.6 & 73.5 & 90.7 & 83.9 & 82.8 & 94.8 & 69.1 & 94.2 & 80.9 & 53.1 & 72.9 & 83.0\\
				PointNet++ \cite{qi2017pointnet++}  & 85.1 & 82.4 & 79.0 & 87.7 & 77.3 & 90.8 & 71.8 & 91.0 & 85.9 & {83.7} & 95.3 & {71.6} & 94.1 & 81.3 & 58.7 & 76.4 & 82.6\\
				SynSpecCNN \cite{yi2017syncspeccnn}  & 84.7 & 81.6 & 81.7 & 81.9 & 75.2 & 90.2 & 74.9 & 93.0 & 86.1 & 84.7 & 95.6 & 66.7 & 92.7 & 81.6 & 60.6 & 82.9 & 82.1\\
				PCNN \cite{atzmon2018point} &  85.1 & 82.4 & 80.1 & 85.5 & 79.5 & 90.8 & 73.2 & 91.3 & 86.0 & 85.0 & 95.7 & 73.2 & 94.8 & 83.3 & 51.0 & 75.0 & 81.8 \\
				SpiderCNN \cite{xu2018spidercnn} & 85.3 & 83.5 & 81.0 & 87.2 & 77.5 & 90.7 & 76.8 & 91.1 & 87.3 & 83.3 & 95.8 & 70.2 & 93.5 & 82.7 & 59.7 & 75.8 & 82.8\\
				PointCNN \cite{li2018pointcnn} & 86.1 & 84.1 & \textbf{86.5} & 86.0 & 80.8 & 90.6 & 79.7 & 92.3 & 88.4 & 85.3 & 96.1 & 77.2 & 95.3 & 84.2 & 64.2 & 80.0 & 83.0\\
				Point2Seq \cite{liu2019point2sequence} & 85.2 & 82.6 & 81.8 & 87.5 & 77.3 & 90.8 & 77.1 & 91.1 & 86.9 & 83.9 & 95.7 & 70.8 & 94.6 & 79.3 & 58.1 & 75.2 & 82.8\\
				RS-CNN \cite{hu2021rscnn} & 86.2 & 83.5 & 84.8 & 88.8 & 79.6 & 91.2 & \textbf{81.1} & 91.6 & 88.4 & 86.0 & 96.0 & 73.7 & 94.1 & 83.4 & 60.5 & 77.7 & 83.6\\
				KPConv \cite{thomas2019kpconv} & \textbf{86.4} & 84.6 & 86.3 & 87.2 & {81.1} & {91.1} & 77.8 & \textbf{92.6} & 88.4 & 82.7 & \textbf{96.2} & \textbf{78.1} & \textbf{95.8} & \textbf{85.4} & \textbf{69.0} & 82.0 & 83.6\\
				3D-GCN \cite{cho2018three} & 85.1 & 83.1 & 84.0 & 86.6 & 77.5 & 90.3 & 74.1 & 90.0 & 86.4 & 83.8 & 95.6 & 66.8 & 94.8 & 81.3 & 59.6 & 75.7 & 82.8\\
				DGCNN \cite{wang2019dynamic}  & 85.2 & 84.0 & 83.4 & 86.7 & 77.8 & 90.6 & 74.7 & {91.2} & 87.5 & 82.8 & 95.7 & 66.3 & 94.9 & 81.1 & 63.5 & 74.5 & 82.6 \\
				PAConv \cite{xu2021paconv} & 86.1 & {84.3} & 85.0 & \textbf{90.4} & 79.7 & 90.6 & {80.8} & 92.0 & \textbf{88.7} & 82.2 &  95.9 & 73.9 & 94.7 & 84.7 & 65.9 & 81.4 & \textbf{84.0}\\
				\hline
				PCT \cite{guo2020pct} & \textbf{86.4} & 85.0 & 82.4 & 89.0 & 81.2 & \textbf{91.9} & 71.5 & 91.3 & 88.1 & \textbf{86.3} & 95.8 & 64.6 & \textbf{95.8} & 83.6 & 62.2 & 77.6 & 83.7\\
				PS-Former (ours) & \textbf{86.4} & \textbf{85.2} & 81.3 & 86.1 & \textbf{81.4} & 91.6 & 72.0 & 92.3 & 88.4 & 85.5 & \textbf{96.2} & 73.5 & 95.3 & 83.7 & 61.7 & 76.9 & 82.9
			\end{tabular}
			}
			\vspace{2mm}
			\caption{\textbf{Part Segmentation Results on ShapeNet:} mIOU is the instance level mean IOU. All results are cited from the original papers.}
	\label{tab:partseg}
\end{table*}

\begin{figure}[!htbp]
\begin{center}
\scalebox{0.84}{
\begin{tabular}{cccc}
\includegraphics[height=0.25\linewidth]{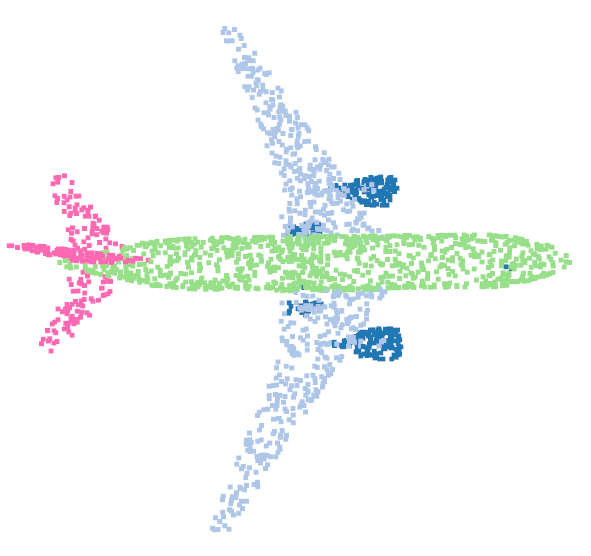}
&
\includegraphics[height=0.25\linewidth]{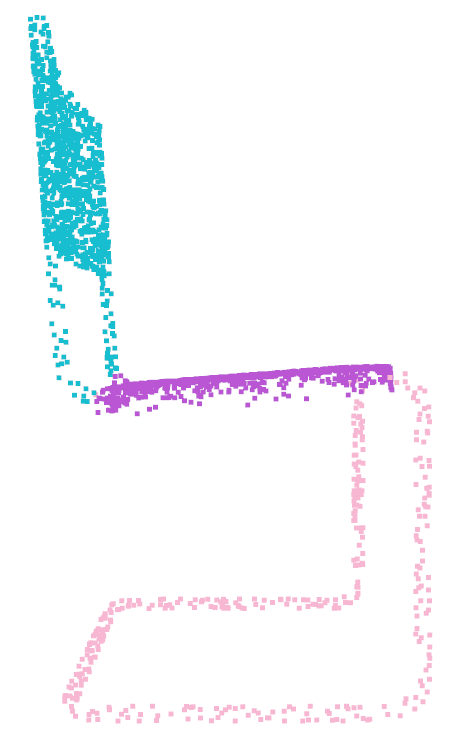}
&
\includegraphics[height=0.25\linewidth]{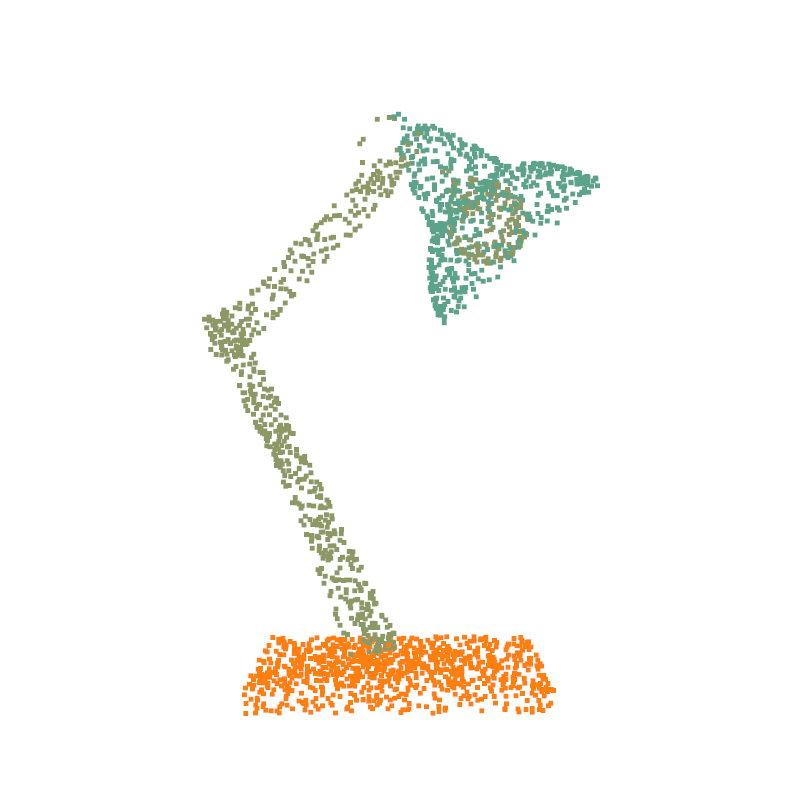}
&
\includegraphics[height=0.25\linewidth]{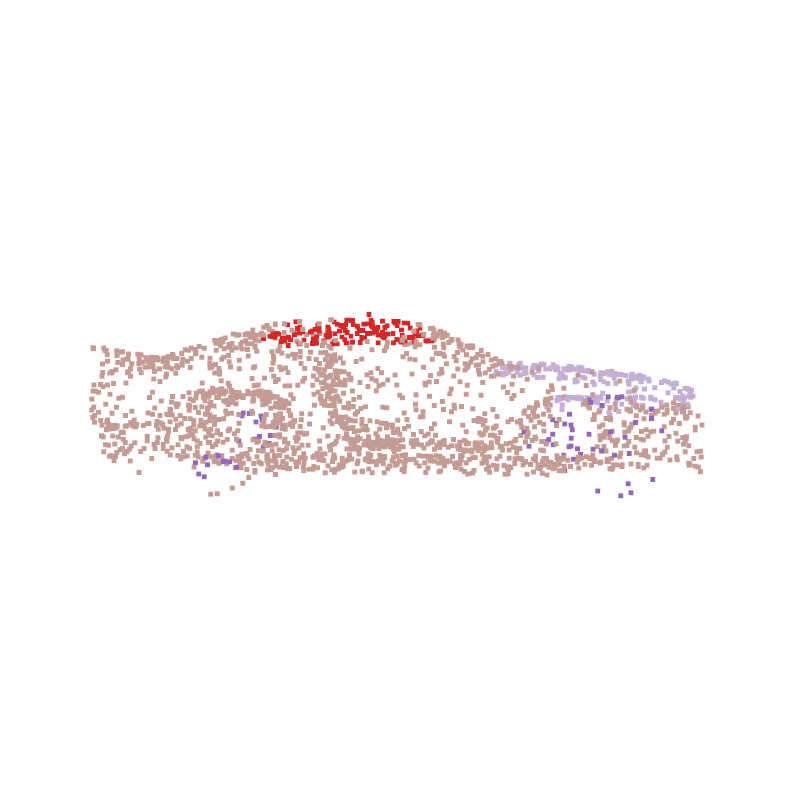}
\\
\includegraphics[height=0.25\linewidth]{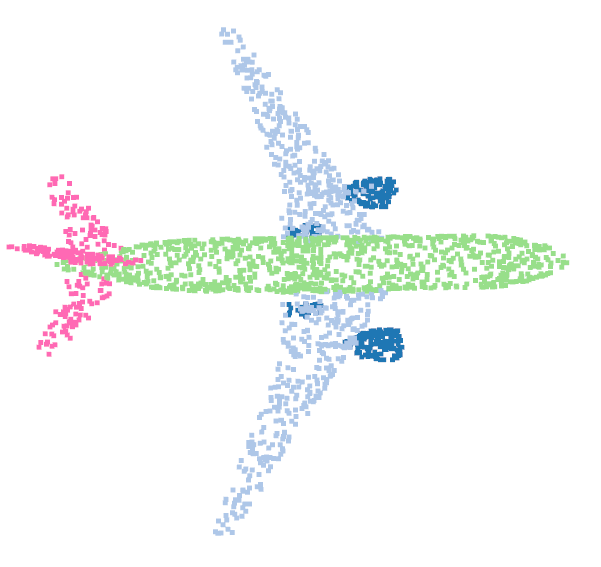}
&
\includegraphics[height=0.25\linewidth]{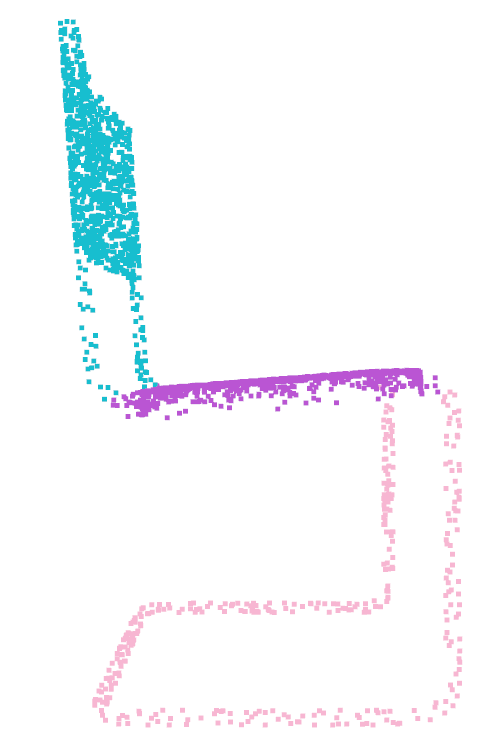}
&
\includegraphics[height=0.25\linewidth]{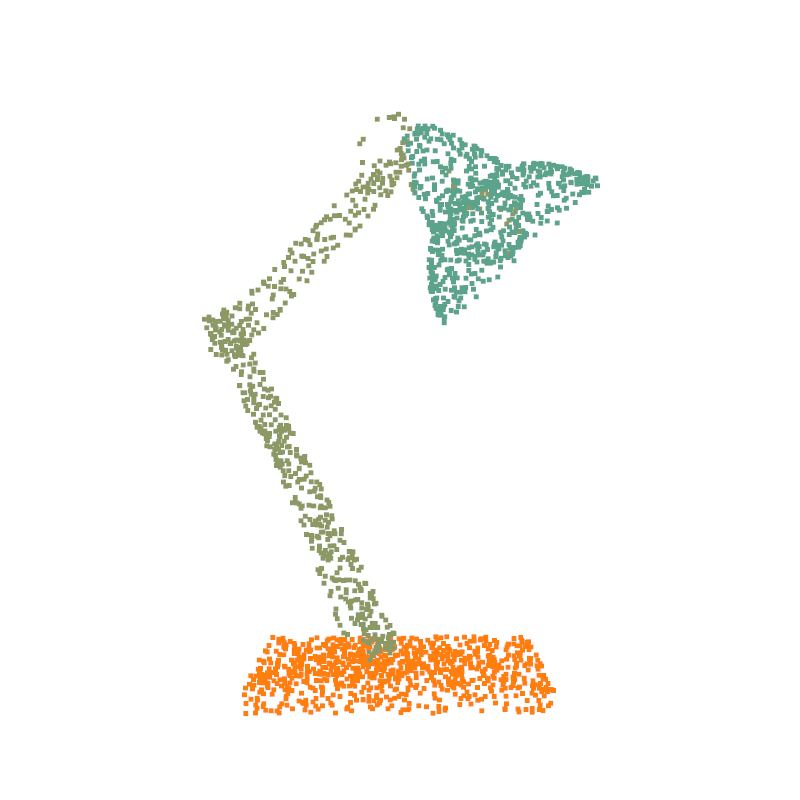}
&
\includegraphics[height=0.25\linewidth]{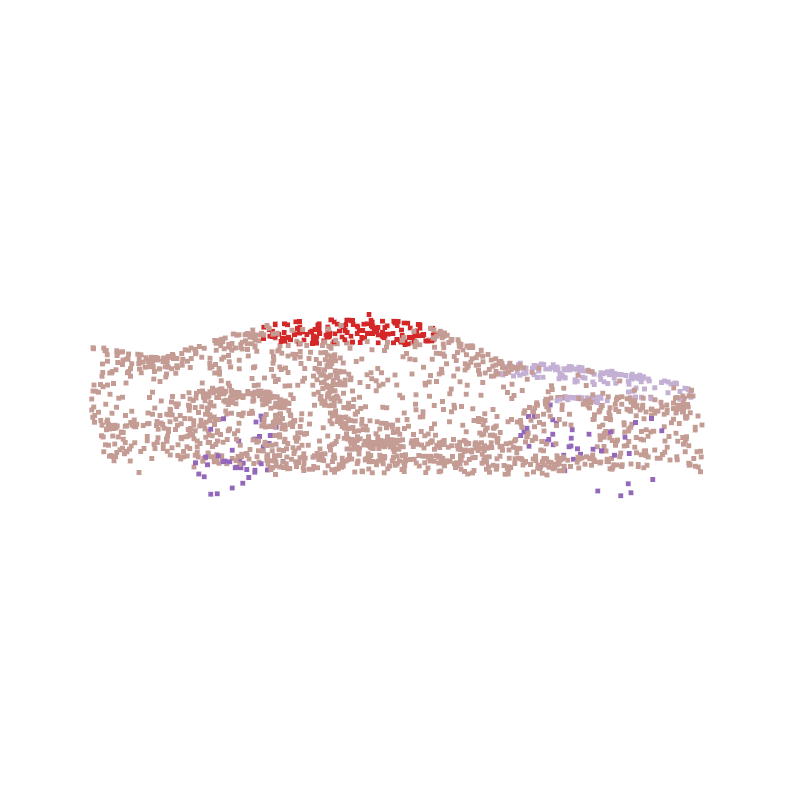}
\end{tabular}
}
\caption{\small \textbf{Visualization on ShapeNet part segmentation:} The first row is the ground truth and the second row is our model's prediction. Each color represents a part label. From left to right are plane, chair, lamp and car respectively.}
\label{fig:partseg-vis}
\end{center}
\end{figure}

\textbf{Results and analysis.} The IoUs of all the test dataset and each class are presented in Table \ref{tab:partseg}. It can be seen that our model achieves competitive results compared to previous methods. Some of the visualization results are presented in Figure \ref{fig:partseg-vis}.

\subsection{Scene Segmentation on S3DIS}

\textbf{Dataset and implementation.} We also use S3DIS\cite{armeni20163d} to evaluate our model on scene segmentation. The dataset contains 6 areas and 271 rooms in total. Each data sample contains 4,096 points with each point assigned with a label from 13 classes. Different from ModelNet40 classification and ShapeNet part segmentation, the input of each point contains 9-dimensional features including coordinate, RGB and normal vector information. Due to this change, our model removes subtraction in the gather operation in the Condensation Layer since we want to keep the color and norm information in each point's structure information while keeping the position information to continue gradually enriching the structure information through the Position-to-Structure Attention Layer. In our training-testing procedure, the training dataset is all the areas except area 5 which leaves area 5 to be the testing dataset. The same optimizer and learning schedule are used as part segmentation which is SGD with the initial learning to 0.1 and Cosine Annealing Scheduler\cite{loshchilov2016sgdr}.

\begin{figure*}[!htp]
\centering
\includegraphics[width=0.8\textwidth]{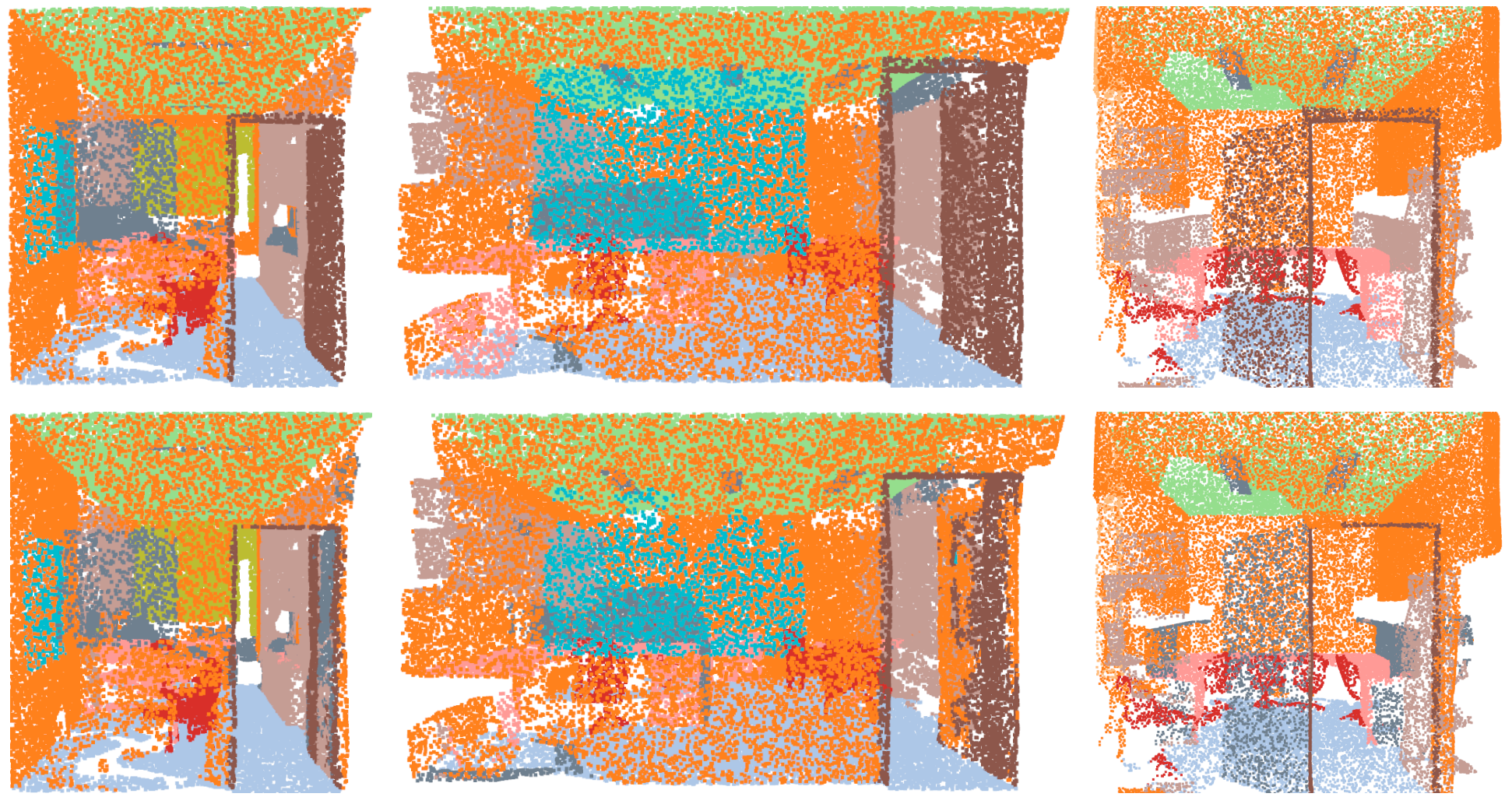}
\caption{\small \textbf{Visualization on S3DIS scene segmentation.} First row: Ground truth. Second row: Our model's prediction.}
\label{fig:semseg-vis}
\vspace{0mm}
\end{figure*}

\textbf{Results and analysis.} The mAcc and mIou of the test dataset are presented in Table \ref{tab:semseg}. 
Note that since our model is trying to decouple the position and structure features and gradually enrich the structure features through the Position-to-Structure Attention Layer, this task which includes RGB and normal vector input as well as position input is not a direct fit for our task. However, it can still benefit from it given by our results. Furthermore, we present several visualization results on Area 5 in Figure \ref{fig:semseg-vis}.

\vspace{0mm}
\begin{table}[!htp]
\centering
\scalebox{0.8}{
\begin{tabular}{l | cc}
\textbf{Model} & mAcc &  mIoU  \\
\hline
PointNet   \cite{qi2017pointnet}
            & 48.98 & 41.09   \\
PCNN \cite{atzmon2018point}
            & 67.01 & 58.27 \\
PointCNN   \cite{li2018pointcnn}
            & 63.86 & 57.26   \\

SPG \cite{landrieu2018large} 
            & 66.50 & 58.04 \\
HPEIN \cite{jiang2019hierarchical}
            & 68.30 & 61.85 \\
DGCNN \cite{wang2019dynamic}
& 84.10 & 56.10 \\
KPConv \cite{thomas2019kpconv}
& 72.8 & 67.1 \\
PointWeb \cite{zhao2019pointweb}
& 66.64 & 60.28 \\
PointTransformer \cite{zhao2021point}
& 76.5 & 70.4 \\
PCT   \cite{guo2020pct}
            & 67.65 & 61.33   \\
PS-Former (ours)
            & 68.57 & 59.36 
\end{tabular}}
\caption{\small Scene Segmentation Results on S3DIS Area 5.}
\label{tab:semseg}
\vspace{0mm}
\end{table}

\section{Ablation Studies}
In this section we explore the effects and impact of the proposed components of the PS-Former---Position-to-Structure Attention, Condensation Layer---and performances of other model variations through ablations evaluated on ModelNet40 classification. Training parameters and testing measures are kept consistent with the description in Section \ref{subsec:modelnet40-experiment}.

\subsection{Position-to-Structure Attention}
\label{subsec:ablation-psa}
We analyze the importance of PS-Former's awareness and distinction between position information and structure information by altering its use and access to such features, as reported in Table \ref{tab:ablation-psa}. We conduct two ablations: 1) removing structure features completely, 2) combining structure features with position features. For the first ablation, the position-only features are fed through vanilla attention. Input dimension is doubled to keep attention dimensions and modeling capability constant. We observe that our model with the proposed Position-to-Structure Attention exhibits a +1.8\% improvement over a position-only attention model. The decrease in performance for the first ablation likely comes from the loss of information on local structures. In the second ablation, the structure features are computed once and concatenated with position features to be passed through classical attention. Compared to the ablation, our proposed method performs +1.1\% better. The results of the two ablations not only suggest the importance of structure features but also of an awareness of the distinction between structure and position.

\begin{table}[!htp]
\centering

\scalebox{1.0}{
\begin{tabular}{l | cc}
\textbf{Configuration} & {Accuracy (\%)}  \\
\hline
{No Structure} & 92.1 \\
{Combined} & 92.8 \\ 
{Position and Structure} & 93.9\\ 
\end{tabular}
}
\caption{\small \textbf{Ablation on Position-to-Structure Attention:} Overall accuracy for classification on ModelNet40\cite{wu20153d} with configurations described in Section \ref{subsec:ablation-psa}.}
\label{tab:ablation-psa}

\end{table}

\begin{table}[!htp]
\centering

\scalebox{1.0}{
\begin{tabular}{l | cc}
\textbf{\# Sampled} & {Accuracy (\%)}  \\
\hline
\textbf{\textsc{1024}} & 93.4 \\
\textbf{\textsc{512}} & 93.9 \\ 
\textbf{\textsc{256}} & 93.6\\ 
\textbf{\textsc{128}} & 93.1\\ 
\end{tabular}
}
\caption{\small \textbf{Ablation on Condensation Layer:} Overall accuracy for classification on ModelNet40\cite{wu20153d} with configurations described in Section \ref{subsec:ablation-condensation}.}
\label{tab:ablation-condensation}

\end{table}




\subsection{Condensation Layer}
\label{subsec:ablation-condensation}
In an effort to demonstrate the Condensation Layer's performance improvements, we conduct ablations on the number of points sampled. As shown in Table \ref{tab:ablation-condensation}, our proposed model with the Condensation Layer configured to 512 points performed better than 1024 points, suggesting the efficacy of our attention-based point sampling in creating information rich local structures. This observation coincides with our hypothesis made in Section \ref{subsec:modelnet40-experiment}. Furthermore, the down-sampling of points provides great relief in terms of computational efficiency because attention is an operation with $\mathcal{O}(n^{2} \cdot d)$ complexity. Through the stated ablations, we verify the utility and desired properties of our proposed Condensation Layer.

\subsection{Multi-Scale Approach}
Previous research has demonstrated a boost in performance gained from multi-scale features that incorporate both global and local structure\cite{qi2017pointnet++}. To investigate its effects regarding our PS-Former, we trained a multi-scale PS-Former and applied it to ModelNet40 for classification. The model is comprised of two branches: a PS-Former that samples 256 points and one that samples 1024 points. The features generated by the two are concatenated together after max pooling and fed through the same classification network used for Section \ref{subsec:modelnet40-experiment}. The model attained a best overall accuracy of 93.6\%. We hypothesize that the multi-scale approach does not provide complementary information, as our Condensation Layer and Position-to-Structure Attention layers already serve as effective extractors of structural feature representation.

\subsection{Cross Attention}
We also conduct ablation studies on using the cross-attention operation to mediate position and structure feature interaction. After the Condensation Layer each point will have two features: position and structure. We use cross-attention to enrich the structure features from the position features. Besides this, we also conduct two experiments using cross-attention from structure to position and two-way interaction. The results are presented in Table \ref{tab:ablation-cross}. It can be seen clearly that with cross-attention from position features to structures, the result is still decent compared to PCT\cite{guo2020pct}. However, the application of cross-attention on structure to position, causes the results to go down. This further demonstrates that the weighted sum mechanism is not suited for capturing the structure features from the position features while the decoupling still works. The decrease in performance with structure enriching position validates our design of position to structure.

\begin{table}[htbp]
\begin{center}
\scalebox{1.0}{
\begin{tabular}{l | cc}
\textbf{Method} & {Accuracy (\%)}  \\
\hline
 (Vanilla cross attention) & \\
pos $\rightarrow$ str & 93.3 \\
str $\rightarrow$ pos & 92.5 \\ 
pos $\rightarrow$ str \& str $\rightarrow$ pos & 92.8\\ 
\hline
Position-to-Structure Attn (ours) & \textbf{93.9}
\end{tabular}
}
\caption{\small \textbf{Ablation on using cross attention to interact between position information and structure information:} Overall accuracy for classification on ModelNet40\cite{wu20153d}}
\label{tab:ablation-cross}
\end{center}
\end{table}

\section{Conclusion}




In this paper, we have presented Position-to-Structure Transformer (PS-Former) for 3D points by designing a learnable Condensation Layer and Position-to-Structure Attention mechanism for effective and efficient point cloud feature extraction. Compared to the existing Transformer-based approaches in this domain, PS-Former is more general with less artificial feature engineering, as well as improved performance.

\noindent {\bf Acknowledgement}
This work is supported by NSF Award IIS-2127544.

{\small
\bibliographystyle{ieee_fullname}
\bibliography{egbib}

\begin{thebibliography}{10}\itemsep=-1pt

\bibitem{armeni20163d}
Iro Armeni, Ozan Sener, Amir~R Zamir, Helen Jiang, Ioannis Brilakis, Martin
  Fischer, and Silvio Savarese.
\newblock 3d semantic parsing of large-scale indoor spaces.
\newblock In {\em Proceedings of the IEEE Conference on Computer Vision and
  Pattern Recognition}, pages 1534--1543, 2016.

\bibitem{atzmon2018point}
Matan Atzmon, Haggai Maron, and Yaron Lipman.
\newblock Point convolutional neural networks by extension operators.
\newblock {\em arXiv preprint arXiv:1803.10091}, 2018.

\bibitem{carion2020end}
Nicolas Carion, Francisco Massa, Gabriel Synnaeve, Nicolas Usunier, Alexander
  Kirillov, and Sergey Zagoruyko.
\newblock End-to-end object detection with transformers.
\newblock In {\em ECCV}, pages 213--229, 2020.

\bibitem{chen2017multi}
Xiaozhi Chen, Huimin Ma, Ji Wan, Bo Li, and Tian Xia.
\newblock Multi-view 3d object detection network for autonomous driving.
\newblock In {\em CVPR}, pages 1907--1915, 2017.

\bibitem{chen2019learning}
Zhiqin Chen and Hao Zhang.
\newblock Learning implicit fields for generative shape modeling.
\newblock In {\em CVPR}, 2019.

\bibitem{cho2018three}
Hyeoncheol Cho and Insung~S Choi.
\newblock Three-dimensionally embedded graph convolutional network (3dgcn) for
  molecule interpretation.
\newblock {\em arXiv preprint arXiv:1811.09794}, 2018.

\bibitem{devlin2018bert}
Jacob Devlin, Ming-Wei Chang, Kenton Lee, and Kristina Toutanova.
\newblock Bert: Pre-training of deep bidirectional transformers for language
  understanding.
\newblock {\em arXiv preprint arXiv:1810.04805}, 2018.

\bibitem{devlin2019bert}
Jacob Devlin, Ming-Wei Chang, Kenton Lee, and Kristina Toutanova.
\newblock Bert: Pre-training of deep bidirectional transformers for language
  understanding.
\newblock In {\em NAACL-HLT}, 2019.

\bibitem{dosovitskiy2021image}
Alexey Dosovitskiy, Lucas Beyer, Alexander Kolesnikov, Dirk Weissenborn,
  Xiaohua Zhai, Thomas Unterthiner, Mostafa Dehghani, Matthias Minderer, Georg
  Heigold, Sylvain Gelly, et~al.
\newblock An image is worth 16x16 words: Transformers for image recognition at
  scale.
\newblock In {\em ICLR}, 2021.

\bibitem{engel2021point}
Nico Engel, Vasileios Belagiannis, and Klaus Dietmayer.
\newblock Point transformer.
\newblock {\em IEEE Access}, 9:134826--134840, 2021.

\bibitem{gkioxari2019mesh}
Georgia Gkioxari, Jitendra Malik, and Justin Johnson.
\newblock Mesh r-cnn.
\newblock In {\em CVPR}, 2019.

\bibitem{goyal2021revisiting}
Ankit Goyal, Hei Law, Bowei Liu, Alejandro Newell, and Jia Deng.
\newblock Revisiting point cloud shape classification with a simple and
  effective baseline.
\newblock {\em arXiv preprint arXiv:2106.05304}, 2021.

\bibitem{guo2020pct}
Meng-Hao Guo, Jun-Xiong Cai, Zheng-Ning Liu, Tai-Jiang Mu, Ralph~R Martin, and
  Shi-Min Hu.
\newblock Pct: Point cloud transformer.
\newblock {\em arXiv preprint arXiv:2012.09688}, 2020.

\bibitem{hu2021rscnn}
Linshu Hu, Mengjiao Qin, Feng Zhang, Zhenhong Du, and Renyi Liu.
\newblock Rscnn: A cnn-based method to enhance low-light remote-sensing images.
\newblock {\em Remote Sensing}, 13(1):62, 2021.

\bibitem{hu2020randla}
Qingyong Hu, Bo Yang, Linhai Xie, Stefano Rosa, Yulan Guo, Zhihua Wang, Niki
  Trigoni, and Andrew Markham.
\newblock Randla-net: Efficient semantic segmentation of large-scale point
  clouds.
\newblock In {\em CVPR}, pages 11108--11117, 2020.

\bibitem{jaritz2019multi}
Maximilian Jaritz, Jiayuan Gu, and Hao Su.
\newblock Multi-view pointnet for 3d scene understanding.
\newblock In {\em Proceedings of the IEEE/CVF International Conference on
  Computer Vision Workshops}, 2019.

\bibitem{jiang2019hierarchical}
Li Jiang, Hengshuang Zhao, Shu Liu, Xiaoyong Shen, Chi-Wing Fu, and Jiaya Jia.
\newblock Hierarchical point-edge interaction network for point cloud semantic
  segmentation.
\newblock In {\em Proceedings of the IEEE/CVF International Conference on
  Computer Vision}, pages 10433--10441, 2019.

\bibitem{kingma2014adam}
Diederik~P Kingma and Jimmy Ba.
\newblock Adam: A method for stochastic optimization.
\newblock {\em arXiv preprint arXiv:1412.6980}, 2014.

\bibitem{klokov2017escape}
Roman Klokov and Victor Lempitsky.
\newblock Escape from cells: Deep kd-networks for the recognition of 3d point
  cloud models.
\newblock In {\em Proceedings of the IEEE International Conference on Computer
  Vision}, pages 863--872, 2017.

\bibitem{lan2019modeling}
Shiyi Lan, Ruichi Yu, Gang Yu, and Larry~S Davis.
\newblock Modeling local geometric structure of 3d point clouds using geo-cnn.
\newblock In {\em Proceedings of the IEEE/CVF Conference on Computer Vision and
  Pattern Recognition}, pages 998--1008, 2019.

\bibitem{landrieu2018large}
Loic Landrieu and Martin Simonovsky.
\newblock Large-scale point cloud semantic segmentation with superpoint graphs.
\newblock In {\em Proceedings of the IEEE conference on computer vision and
  pattern recognition}, pages 4558--4567, 2018.

\bibitem{lecun1989backpropagation}
Y. LeCun, B. Boser, J.~S. Denker, D. Henderson, R.E. Howard, W. Hubbard, and
  L.D. Jackel.
\newblock {Backpropagation applied to handwritten zip code recognition}.
\newblock {\em Neural Computation}, 1989.

\bibitem{li2018so}
Jiaxin Li, Ben~M Chen, and Gim~Hee Lee.
\newblock So-net: Self-organizing network for point cloud analysis.
\newblock In {\em Proceedings of the IEEE conference on computer vision and
  pattern recognition}, pages 9397--9406, 2018.

\bibitem{li2018pointcnn}
Yangyan Li, Rui Bu, Mingchao Sun, Wei Wu, Xinhan Di, and Baoquan Chen.
\newblock Pointcnn: Convolution on x-transformed points.
\newblock {\em Advances in neural information processing systems}, 31:820--830,
  2018.

\bibitem{liu2019point2sequence}
Xinhai Liu, Zhizhong Han, Yu-Shen Liu, and Matthias Zwicker.
\newblock Point2sequence: Learning the shape representation of 3d point clouds
  with an attention-based sequence to sequence network.
\newblock In {\em AAAI}, pages 8778--8785, 2019.

\bibitem{loshchilov2016sgdr}
Ilya Loshchilov and Frank Hutter.
\newblock Sgdr: Stochastic gradient descent with warm restarts.
\newblock {\em arXiv preprint arXiv:1608.03983}, 2016.

\bibitem{mao2019interpolated}
Jiageng Mao, Xiaogang Wang, and Hongsheng Li.
\newblock Interpolated convolutional networks for 3d point cloud understanding.
\newblock In {\em Proceedings of the IEEE/CVF International Conference on
  Computer Vision}, pages 1578--1587, 2019.

\bibitem{maturana2015voxnet}
Daniel Maturana and Sebastian Scherer.
\newblock Voxnet: A 3d convolutional neural network for real-time object
  recognition.
\newblock In {\em 2015 IEEE/RSJ International Conference on Intelligent Robots
  and Systems (IROS)}, pages 922--928. IEEE, 2015.

\bibitem{mazur2021cloud}
Kirill Mazur and Victor Lempitsky.
\newblock Cloud transformers: A universal approach to point cloud processing
  tasks.
\newblock In {\em Proceedings of the IEEE/CVF International Conference on
  Computer Vision}, pages 10715--10724, 2021.

\bibitem{meyer2019lasernet}
Gregory~P Meyer, Ankit Laddha, Eric Kee, Carlos Vallespi-Gonzalez, and Carl~K
  Wellington.
\newblock Lasernet: An efficient probabilistic 3d object detector for
  autonomous driving.
\newblock In {\em CVPR}, pages 12677--12686, 2019.

\bibitem{mo2019structurenet}
Kaichun Mo, Paul Guerrero, Li Yi, Hao Su, Peter Wonka, Niloy~J Mitra, and
  Leonidas~J Guibas.
\newblock Structurenet: hierarchical graph networks for 3d shape generation.
\newblock {\em ACM Transactions on Graphics}, 38(6):1--19, 2019.

\bibitem{qi2017pointnet}
Charles~R Qi, Hao Su, Kaichun Mo, and Leonidas~J Guibas.
\newblock Pointnet: Deep learning on point sets for 3d classification and
  segmentation.
\newblock In {\em Proceedings of the IEEE conference on computer vision and
  pattern recognition}, pages 652--660, 2017.

\bibitem{qi2016volumetric}
Charles~R Qi, Hao Su, Matthias Nie{\ss}ner, Angela Dai, Mengyuan Yan, and
  Leonidas~J Guibas.
\newblock Volumetric and multi-view cnns for object classification on 3d data.
\newblock In {\em Proceedings of the IEEE conference on computer vision and
  pattern recognition}, pages 5648--5656, 2016.

\bibitem{qi2017pointnet++}
Charles~R Qi, Li Yi, Hao Su, and Leonidas~J Guibas.
\newblock Pointnet++: Deep hierarchical feature learning on point sets in a
  metric space.
\newblock {\em arXiv preprint arXiv:1706.02413}, 2017.

\bibitem{ran2021learning}
Haoxi Ran, Wei Zhuo, Jun Liu, and Li Lu.
\newblock Learning inner-group relations on point clouds.
\newblock In {\em Proceedings of the IEEE/CVF International Conference on
  Computer Vision}, pages 15477--15487, 2021.

\bibitem{shaw2018self}
Peter Shaw, Jakob Uszkoreit, and Ashish Vaswani.
\newblock Self-attention with relative position representations.
\newblock In {\em Proceedings of the 2018 Conference of the North American
  Chapter of the Association for Computational Linguistics: Human Language
  Technologies, Volume 2 (Short Papers)}, 2018.

\bibitem{su2015multi}
Hang Su, Subhransu Maji, Evangelos Kalogerakis, and Erik Learned-Miller.
\newblock Multi-view convolutional neural networks for 3d shape recognition.
\newblock In {\em Proceedings of the IEEE international conference on computer
  vision}, pages 945--953, 2015.

\bibitem{thomas2019kpconv}
Hugues Thomas, Charles~R Qi, Jean-Emmanuel Deschaud, Beatriz Marcotegui,
  Fran{\c{c}}ois Goulette, and Leonidas~J Guibas.
\newblock Kpconv: Flexible and deformable convolution for point clouds.
\newblock In {\em Proceedings of the IEEE/CVF International Conference on
  Computer Vision}, pages 6411--6420, 2019.

\bibitem{vaswani2017attention}
Ashish Vaswani, Noam Shazeer, Niki Parmar, Jakob Uszkoreit, Llion Jones,
  Aidan~N Gomez, {\L}ukasz Kaiser, and Illia Polosukhin.
\newblock Attention is all you need.
\newblock In {\em Advances in neural information processing systems}, pages
  5998--6008, 2017.

\bibitem{wang2018local}
Chu Wang, Babak Samari, and Kaleem Siddiqi.
\newblock Local spectral graph convolution for point set feature learning.
\newblock In {\em Proceedings of the European conference on computer vision
  (ECCV)}, pages 52--66, 2018.

\bibitem{wang2019dynamic}
Yue Wang, Yongbin Sun, Ziwei Liu, Sanjay~E Sarma, Michael~M Bronstein, and
  Justin~M Solomon.
\newblock Dynamic graph cnn for learning on point clouds.
\newblock {\em Acm Transactions On Graphics (tog)}, 38(5):1--12, 2019.

\bibitem{wu2017marrnet}
Jiajun Wu, Yifan Wang, Tianfan Xue, Xingyuan Sun, Bill Freeman, and Josh
  Tenenbaum.
\newblock Marrnet: 3d shape reconstruction via 2.5 d sketches.
\newblock In {\em Advances in Neural Information Processing Systems}, 2017.

\bibitem{wu2019pointconv}
Wenxuan Wu, Zhongang Qi, and Li Fuxin.
\newblock Pointconv: Deep convolutional networks on 3d point clouds.
\newblock In {\em Proceedings of the IEEE/CVF Conference on Computer Vision and
  Pattern Recognition}, pages 9621--9630, 2019.

\bibitem{wu20153d}
Zhirong Wu, Shuran Song, Aditya Khosla, Fisher Yu, Linguang Zhang, Xiaoou Tang,
  and Jianxiong Xiao.
\newblock 3d shapenets: A deep representation for volumetric shapes.
\newblock In {\em Proceedings of the IEEE conference on computer vision and
  pattern recognition}, pages 1912--1920, 2015.

\bibitem{xiang2021walk}
Tiange Xiang, Chaoyi Zhang, Yang Song, Jianhui Yu, and Weidong Cai.
\newblock Walk in the cloud: Learning curves for point clouds shape analysis.
\newblock {\em ICCV}, 2021.

\bibitem{xie2018attentional}
Saining Xie, Sainan Liu, Zeyu Chen, and Zhuowen Tu.
\newblock Attentional shapecontextnet for point cloud recognition.
\newblock In {\em Proceedings of the IEEE Conference on Computer Vision and
  Pattern Recognition}, pages 4606--4615, 2018.

\bibitem{xu2021paconv}
Mutian Xu, Runyu Ding, Hengshuang Zhao, and Xiaojuan Qi.
\newblock Paconv: Position adaptive convolution with dynamic kernel assembling
  on point clouds.
\newblock In {\em Proceedings of the IEEE/CVF Conference on Computer Vision and
  Pattern Recognition}, pages 3173--3182, 2021.

\bibitem{xu2020grid}
Qiangeng Xu, Xudong Sun, Cho-Ying Wu, Panqu Wang, and Ulrich Neumann.
\newblock Grid-gcn for fast and scalable point cloud learning.
\newblock In {\em Proceedings of the IEEE/CVF Conference on Computer Vision and
  Pattern Recognition}, pages 5661--5670, 2020.

\bibitem{xu2018spidercnn}
Yifan Xu, Tianqi Fan, Mingye Xu, Long Zeng, and Yu Qiao.
\newblock Spidercnn: Deep learning on point sets with parameterized
  convolutional filters.
\newblock In {\em Proceedings of the European Conference on Computer Vision
  (ECCV)}, pages 87--102, 2018.

\bibitem{yang2019modeling}
Jiancheng Yang, Qiang Zhang, Bingbing Ni, Linguo Li, Jinxian Liu, Mengdie Zhou,
  and Qi Tian.
\newblock Modeling point clouds with self-attention and gumbel subset sampling.
\newblock In {\em Proceedings of the IEEE/CVF Conference on Computer Vision and
  Pattern Recognition}, pages 3323--3332, 2019.

\bibitem{yi2016scalable}
Li Yi, Vladimir~G Kim, Duygu Ceylan, I-Chao Shen, Mengyan Yan, Hao Su, Cewu Lu,
  Qixing Huang, Alla Sheffer, and Leonidas Guibas.
\newblock A scalable active framework for region annotation in 3d shape
  collections.
\newblock {\em ACM Transactions on Graphics (ToG)}, 35(6):1--12, 2016.

\bibitem{yi2017syncspeccnn}
Li Yi, Hao Su, Xingwen Guo, and Leonidas~J Guibas.
\newblock Syncspeccnn: Synchronized spectral cnn for 3d shape segmentation.
\newblock In {\em Proceedings of the IEEE Conference on Computer Vision and
  Pattern Recognition}, pages 2282--2290, 2017.

\bibitem{zhao2019pointweb}
Hengshuang Zhao, Li Jiang, Chi-Wing Fu, and Jiaya Jia.
\newblock {PointWeb}: Enhancing local neighborhood features for point cloud
  processing.
\newblock In {\em CVPR}, 2019.

\bibitem{zhao2021point}
Hengshuang Zhao, Li Jiang, Jiaya Jia, Philip~HS Torr, and Vladlen Koltun.
\newblock Point transformer.
\newblock In {\em Proceedings of the IEEE/CVF International Conference on
  Computer Vision}, pages 16259--16268, 2021.

\end{thebibliography}
}

\end{document}